\documentclass[runningheads]{llncs}

\usepackage{eccv}

\usepackage{eccvabbrv}

\usepackage{graphicx}
\usepackage{booktabs}

\usepackage[accsupp]{axessibility}  

\usepackage{hyperref}

\usepackage{orcidlink}

\begin{document}

\title{Linking in Style: Understanding learned features in deep learning models} 

\titlerunning{Linking in Style}

\author{Maren H. Wehrheim\inst{1, 2}\orcidlink{0000-0003-3197-9947} \and
Pamela Osuna-Vargas\inst{1, 2}\orcidlink{0009-0005-9154-7515} \and
Matthias Kaschube\inst{1, 2}\orcidlink{0000-0002-5145-7487}}

\authorrunning{~Wehrheim et al.}

\institute{Frankfurt Institute for Advanced Studies (FIAS), Frankfurt, Germany
 \and
Department of Computer Science and Mathematics, Goethe University Frankfurt, Frankfurt, Germany\\
\email{\{wehrheim, osuna, kaschube\}@fias.uni-frankfurt.de}}

\maketitle

\begin{center}
    \centering
    \includegraphics[width=\textwidth]{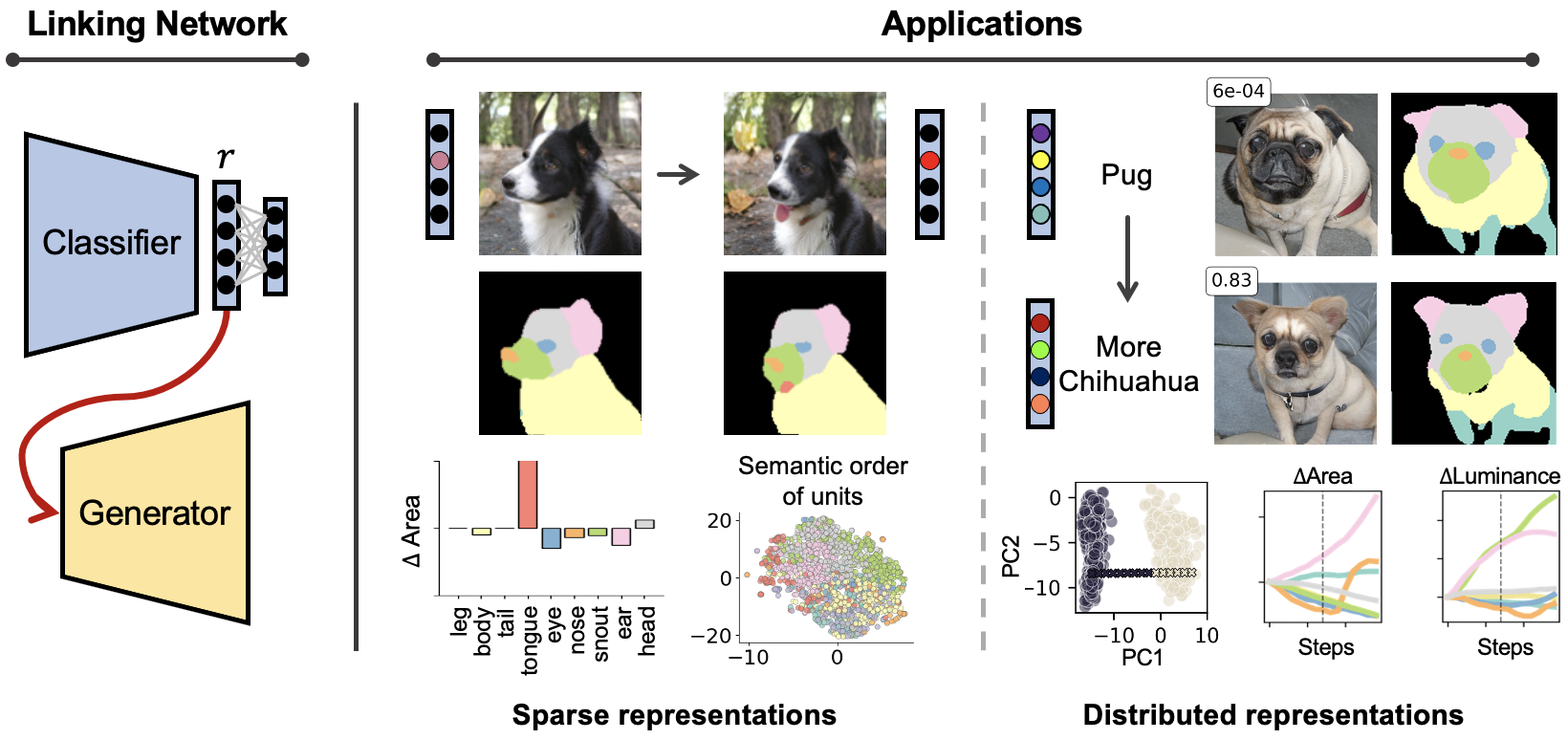}
    \captionof{figure}{\textbf{Visualization and systematic quantification of a classifier's learned representations}. \textbf{Left:} We introduce a linking network (red arrow) that links an activation pattern $r \in R$ in the penultimate layer of a classifier to the latent space of StyleGAN-XL \cite{sauer_stylegan-xl_2022}, thereby visualizing the representations learned by the classifier. Building on these visualizations, we propose a pipeline to automatically and objectively analyze a large number of learned representations in $R$ by evaluating the changes between images caused by perturbations in $R$. We show two applications of how our method can be used to understand learned features in deep learning models. \textbf{Middle:} We systematically 'tune' the activations of all units in $R$ separately to obtain a comprehensive overview across sparsely encoded representations across thousands of units. \textbf{Right:} The linking network can visualize counterfactual examples and our quantification pipeline reveals trajectories that provide insights into learned concepts relevant for the classifier's decision.}
    \label{fig:fig1}
\end{center}%

\begin{abstract}

Convolutional neural networks (CNNs) learn abstract features to perform object classification, but understanding these features remains challenging due to difficult-to-interpret results or high computational costs. We propose an automatic method to visualize and systematically analyze learned features in CNNs. Specifically, we introduce a linking network that maps the penultimate layer of a pre-trained classifier to the latent space of a generative model (StyleGAN-XL), thereby enabling an interpretable, human-friendly visualization of the classifier's representations.  Our findings indicate a congruent semantic order in both spaces, enabling a direct linear mapping between them. Training the linking network is computationally inexpensive and decoupled from training both the GAN and the classifier. We introduce an automatic pipeline that utilizes such GAN-based visualizations to quantify learned representations by analyzing activation changes in the classifier in the image domain. This quantification allows us to systematically study the learned representations in several thousand units simultaneously and to extract and visualize units selective for specific semantic concepts. Further, we illustrate how our method can be used to quantify and interpret the classifier's decision boundary using counterfactual examples. Overall, our method offers systematic and objective perspectives on learned abstract representations in CNNs. \url{https://github.com/kaschube-lab/LinkingInStyle.git}

\end{abstract}    
\section{Introduction}
\label{sec:intro}
Deep learning models learn abstract concepts in their hidden layers when trained to perform a task. However, the models do not provide plausible explanations when they fail, thereby hampering their trustworthiness. Unraveling the learned concepts that influence a classifier's decisions can reveal inherent biases \cite{kim_interpretability_2018, goyal_inductive_2022} or identify failures in these models \cite{varoquaux_machine_2022, oakden-rayner_hidden_2020}.

Recent work has focused on interpreting deep learning models' behavior by explaining their weights, units, subnetworks, or latent representations \cite{rauker_toward_2023}. Individual units in deep neural networks (DNNs) have been shown to be selective for single human-interpretable concepts such as faces, food, textures, or even multimodal concepts \cite{zhou_object_2015, goh_multimodal_nodate, baek_face_2021, radford_learning_2017, bau_identifying_2018, mahendran_understanding_2014, zeiler_visualizing_2014}. It has since been an open debate whether DNNs learn disentangled, sparse representations in individual units or whether representations are distributed across many units, a phenomenon often referred to as feature superposition \cite{hassabis_neuroscience-inspired_2017, greff_binding_2020, elhage_toy_nodate} and hypothesized to contribute to adversarial vulnerability \cite{gilmer_adversarial_2018, engstrom_adversarial_2019}.

Recent efforts have also been dedicated to understanding the representations that form the decision boundaries in DNNs trained for visual object classification \cite{he_decision_2018, karimi_decision_2020, karimi_characterizing_2020, somepalli_can_2022}. Counterfactual explanations present an empirical perspective for interpreting how deep learning models make decisions. Consider an instance of a given class (e.g., an image of a dog), a \textit{counterfactual example} represents a slightly altered version of that instance such that the classifier predicts a target class (e.g., cat). Crucially, the change in the original instance should be minimal and human-interpretable to be effective. This excludes adversarial examples \cite{zhang_adversarial_2020, yuan_adversarial_2019, buckner_understanding_2020}, where small pixel perturbations change the prediction but remain unrecognized by humans. 

For computer vision applications, explainability methods greatly benefit from providing human-comprehensible visualizations of single examples. However, human visual inspection is inherently subjective and only possible for a few features, prohibiting an unbiased and systematic evaluation of the high-dimensional representations in CNNs. Studying all potential configurations of representations poses an intricate combinatorial challenge, hence visual inspection soon becomes infeasible and cannot provide a comprehensive and objective understanding of learned features in hidden layers.  

Generative adversarial networks (GANs) \cite{goodfellow_generative_2014} are characterized by a latent space that is continuous and semantically structured, enabling the visualization of feature representations, including counterfactual examples \cite{lang_explaining_2021, singla_explaining_2022}. 
However, as GANs were usually trained with a single data category to ensure the generation of high-quality images, their ability to visualize learned representations in classification models required extensive (re-)training and remained infeasible for multi-class categorization problems. Only recently, the StyleGAN-XL \cite{sauer_stylegan-xl_2022} allows to generate images of all ImageNet classes from a single latent space.

In this work, we present a broadly applicable method to objectively and systematically analyze features encoded in the penultimate layer of CNNs trained for object classification. This is achieved in two steps: Firstly we establish an efficient feature visualization tool based on a pre-trained StyleGAN-XL that can be flexibly linked to various pre-trained CNNs (overcoming extensive (re-)training strategies of several previous studies). Specifically, we introduce a linking network that connects the penultimate layer, here termed \textit{representation space}, of a CNN to the latent space of a pre-trained StyleGAN-XL. Linking these two spaces allows us to visualize arbitrary feature dimensions in the classifier. Secondly, we establish methods for an automatic assessment of learned features in the classifier's representation space using unsupervised tracking methods \cite{revaud_pump_2022} and few-shot image segmentation \cite{tritrong_repurposing_2021}. We envision our pipeline to offer novel research applications and show examples in Sec. \ref{sec:results}. First, we analyze and quantify the features encoded in each of the several thousand units of the penultimate layer to build summary statistics of a classifier's learned concepts. This also enables us to reveal class-relevant units encoding human-interpretable features, shedding new light on the recurring question of whether features are represented in individual units in a rather disentangled or superimposed fashion. Second, we probe the classifier's decision boundary to identify and interpret the most relevant features underlying classification. 
Our contributions are as follows:
\begin{itemize}
    \item A simple and easy-to-train linking network to visualize learned representations in CNNs.
    \item An automated pipeline to quantify these high-dimensional representations enabling their systematic analysis and objective characterization via summary statistics.
    \item  We highlight two applications of our method: i) to reveal learned abstract concepts in single units and ii) to examine a classifier's decision boundaries. 
\end{itemize}
\section{Related Work}
\label{sec:related_work}

Explainable AI (XAI) aims to provide human-understandable explanations for the features learned and decisions made by an AI system. In this context, some research argues for the relevance of sparse representations to encode abstract features within single units \cite{dhamdhere_how_2018, dalvi_neurox_2019, ghorbani_neuron_2020, mu_compositional_2020, bau_understanding_2020, amjad_understanding_2022, lundstrom_rigorous_2022}, others highlight the importance of distributed highly robust representations \cite{donnelly_interpretability_2019, leavitt_selectivity_2020, morcos_importance_2018}. A variety of methods that explain the learned representations in pre-trained models exist \cite{alicioglu_survey_2022, rauker_toward_2023}, including GradCAM \cite{selvaraju_grad-cam_2020}, DeepLIFT \cite{shrikumar_learning_2019} or LIME \cite{ribeiro_why_2016}. These methods usually visualize single features or saliency maps but do not quantify the \textit{what} and \textit{how}, e.g., larger eyes or different color, without additional user input.  

GANs generate near photorealistic images \cite{brock_large_2019, karras_analyzing_2020, karras_style-based_2019, karras_training_2020, goodfellow_generative_2014, kang_scaling_2023} and manipulating their latent code smoothly alters features of the generated image \cite{harkonen_ganspace_2020, jahanian_steerability_2020, voynov_unsupervised_2020, shen_closed-form_2021, plumerault_controlling_2020, shen_interpreting_2020, hou_guidedstyle_2022}. Recent work shows that semantic concepts in GANs allow to generate image segmentation masks, using, for example, unsupervised clustering \cite{pakhomov_segmentation_2021, xu_extracting_2022}, few-shot learning \cite{tritrong_repurposing_2021}, or self-supervised contrastive approaches \cite{manerikar_self-supervised_2023}. Representations of (pre-trained) classifiers have previously been used to guide the generative process or to build meaningful latent spaces \cite{casanova_instance-conditioned_2021, bordes_high_2022, lang_explaining_2021, singla_explaining_2022}. In recent work, GANs have been used to visualize changes in single attributes for counterfactual examples of a classifier \cite{lang_explaining_2021, singla_explaining_2022}. Lang et al. \cite{lang_explaining_2021} incorporate a GAN in the training procedure of the classifier and then extract user-defined attributes that change the classifier's output. However, this approach is computationally expensive and does not allow to interpret single units in the classifier. 

Previous work on GAN-based image editing or counterfactual explanations in computer vision often focuses on visualizing learned representations or relies on user input to quantify the learned concepts \cite{guidotti_counterfactual_2022, mothilal_explaining_2020, goyal_counterfactual_2019, harkonen_ganspace_2020, jahanian_steerability_2020, voynov_unsupervised_2020, shen_closed-form_2021, plumerault_controlling_2020, shen_interpreting_2020}. 
\cite{joshi_semantic_2019} and \cite{ vedaldi_semanticadv_2020} define a set of attributes a priori such that a manipulation induces a change in the predicted category. Other methods rely on text guidance to generate difficult images for the classifier \cite{prabhu_lance_2023} or identify the classifier's sensitivity to certain features \cite{luo_zero-shot_2023}. However, methods that enable an objective and comprehensive study of learned(class-relevant) features using photorealistic visualizations and flexibly allowing for an analysis of any combination of units are still lacking.
\section{Method}
\label{sec:methods}
Next, we describe our approach for uncovering learned representations in a classifier. The core of our method is a linking network that learns a mapping between the classifier and the latent space in StyleGAN-XL (Sec. \ref{sec:linking-network}). We then introduce a pipeline to automatically analyze learned concepts (Sec. \ref{sec:analysis-pipeline}). Finally, we propose different applications, demonstrating how our method can be used to understand single units as well as distributed representations relevant for a classifier's decision (Sec. \ref{sec:methods-applications}).

\subsection{Linking network} 
\label{sec:linking-network} 
\begin{figure}[h]
  \centering
  \includegraphics[width=\textwidth]{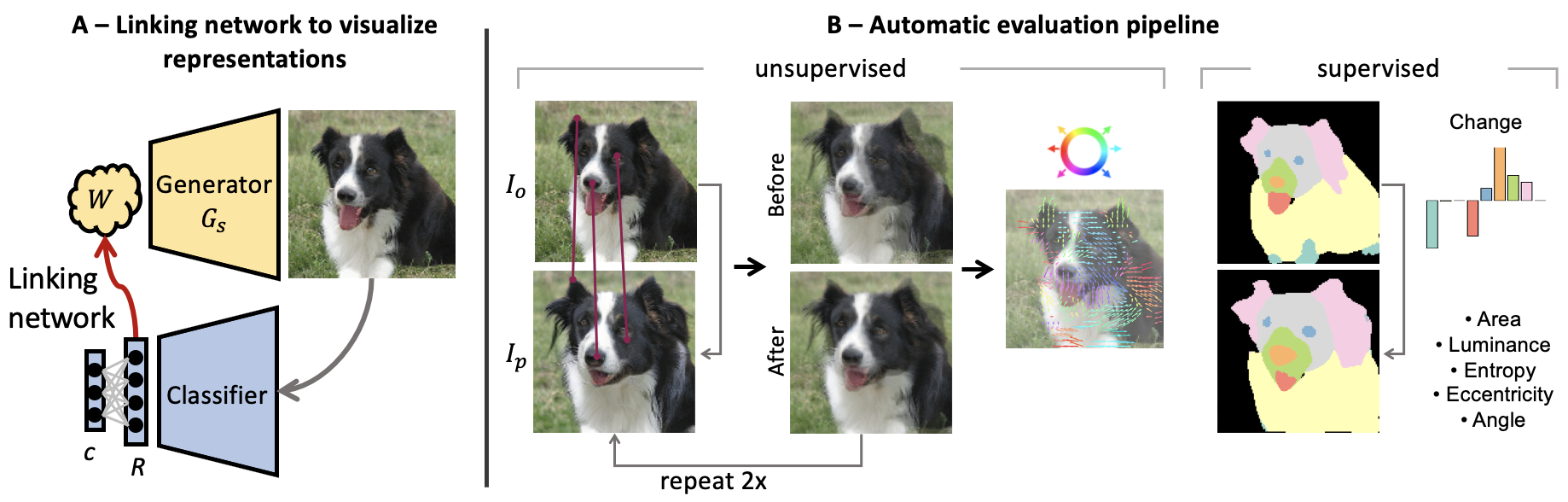}
  \caption{\textbf{Visualizing and quantifying learned features in CNNs}. \textbf{A)} The generator $G_s$ generates an image $I$ from a given $w \in W$. $I$ is input to the classifier from which the corresponding activation vector $r \in R$ is extracted. Using a set of ($w$, $r$)-pairs, we train a linking network (red arrow) to create a link between the classifier and the GAN. 
  We then perturb the activation pattern $r$ to visualize learned representations in $R$ using the GAN.
  \textbf{B)} Automatic quantification of semantic concepts. 
  Left (unsupervised): We introduce an unsupervised method to find matching points between images $I_o$ and $I_p$. First, we use PUMP \cite{revaud_pump_2022} to compute an affine transformation and align the two images to remove global changes such as translation or zoom (center, top: non-aligned images, bottom: aligned images). We then compute PUMP again to find local changes not accounted for by the affine transform and compute the vector field to visualize the changes. Right (supervised): We compute the segmentation mask for each image separately following  \cite{tritrong_repurposing_2021}. Then, we quantify each semantic label in an image according to different evaluation metrics: area (shown here), luminance, entropy, eccentricity, and angle. For each metric, we compute the change induced by a perturbation in $R$.}
  \label{fig:fig2}
\end{figure}

We introduce a linking network that establishes a connection between the classifier and the GAN to visualize learned representations in the classifier (Fig. \ref{fig:fig2} red arrow). We utilize the recently proposed pre-trained StyleGAN-XL \cite{sauer_stylegan-xl_2022}, as it produces high-quality images and learns a single latent code across all classes without extra class-conditional input in the higher layers of the generator. For a given style code $w \in W$, corresponding to a non-linear combination of a class embedding vector $c$ and a random latent vector $z$, the generator $G_s$ creates an image $I$. As we are interested in studying the internal representations of CNNs, we input the generated image into the classifier and extract the respective activation pattern $r$ in the penultimate layer, which we call the representation layer $R$. Following this procedure, we generate 5,000 training instances of ($w$, $r$)-pairs for each class and train a network that maps the activations into $W$: 
\begin{equation}
    \Tilde{w} = f(r),
\end{equation}
where $\Tilde{w}$ is the predicted $w$, $f$ is the linking network and $r \in R$ is a specific activation vector. Thus, the linking network acts as a bridge between the classifier and the generative model and offers the possibility to perform a full cycle, that is $w \to I \to r \to \Tilde{w} \to \Tilde{I}$. In the simplest scenario, we use a linear regression model to fit a linear mapping between the two spaces based on the least-square distance (see Supplement for more complex linking networks). 


\subsection{Analyzing learned representations}
\label{sec:analysis-pipeline} 
CNNs learn abstract semantic concepts like eyes or faces. Whereas humans easily visually detect abstract concepts, quantifying them is challenging. In addition to visualizing examples of learned representations, we here also introduce two methods to objectively and systematically quantify semantic concepts learned by the classifier to increase the explainability of CNNs. Specifically, we introduce an unsupervised method that aids visual inspection and quantifies regional features. Additionally, we propose a supervised method to facilitate the interpretation of learned features using image segmentation (area, luminance, entropy, angle, eccentricity). The features analyzed by these two methods are revealed by comparing an original image $I_o$ to an image $I_p$ generated after perturbing $I_o$'s representation $r$. 

\subsubsection{Unsupervised local descriptors matching}
\label{sec:analysis-pipeline-unsupervised} 
We use an unsupervised motion tracking pipeline to reveal learned features by analyzing differences across images associated with perturbations in the classifier's activation space (Fig. \ref{fig:fig2}B unsupervised). First, we use PUMP \cite{revaud_pump_2022}, an unsupervised method that finds pixel correspondences between $I_o$ and $I_p$, to compute an affine transformation. We then apply the affine transformation to $I_p$, hence minimizing the effect of global displacement (rotation, scaling, etc.). Finally, we again compute PUMP to find the set of dense correspondences and visualize the vector field of local changes.

\subsubsection{Semantic concept quantification through image segmentation}
We hypothesize that a classification model learns abstract semantic concepts (e.g., ear shape, fur type) to differentiate between classes. To capture changes in such semantic concepts, we adopt a few-shot image segmentation approach (Fig. \ref{fig:fig2}B supervised) based on \cite{tritrong_repurposing_2021}. This method leverages the intermediate activation output of the generator $G_s$ to learn an image segmentation model from few labeled examples only. The few-shot nature and generalizability across classes (see Supplement) of this approach reduce the workload required to produce annotations, thus allowing for more detailed labels. To reduce the computational cost we here use only every second layer of $G_s$ and additionally downsample the output to $128\times128$ pixels (instead of $256\times256$). We label five images per class for a subset of classes and train a segmentation model with three convolutional layers for 100 epochs. For any generated image, we can then compute different metrics to quantify the segmented parts, each of which represents a semantic concept (e.g. eyes, tongue). First, we extract the area, luminance, and entropy (smoothness) for each identified segment. Finally, to study its shape and orientation, we fit an ellipse to each segmented component and compute the eccentricity and angle. In total, we characterize each image by 45 measures (5 metrics, 9 semantic labels). Moreover, to test whether a perturbation in $r$ affects only one or many semantic concepts, we define the label sparsity $s$ according to \cite{hoyer_non-negative_nodate} 
\begin{equation}
\label{eq:sparsity}
s(x) = \frac{\sqrt{k} - \frac{\|x\|_1}{\|x\|_2}}{\sqrt{k}-1},
\end{equation}
where $x \in \mathbb{R}^k$ is the change in label vector induced by a perturbation in $r$ and $k$ represents the label vector's dimension (here, $k=9$ considering a metric separately (e.g., area), or $k=45$ for the complete set of measures). If $s$ is close to one, the label vector is sparse, whereas values approaching zero describe distributed representations.

\subsection{Applications}
\label{sec:methods-applications}
We propose several applications to systematically analyze the learned features in the classifier's representation space using our linking network and automatic quantification pipeline.
First, we analyze the representations learned by single units. To this end, we map an image $I_o$ into the classifier and extract the activation $r \in R$. We then alter $r$ of a unit of interest, map the perturbed activation $r_p$ into $W$, and generate an image $I_p$. We can continuously visualize a whole trajectory of the representations encoded in a single unit by linearly altering that unit's activation (between the empirical minimum and maximum) and generating the corresponding sequence of images. We repeat this process for all units in $R$ and quantify the encoded representations using the changes along the image sequence to compute summary statistics across several thousand units and to identify units with certain properties.

Second, we analyze distributed representations that form the classifier's decision boundary. Specifically, we find counterfactual explanations by changing $r$ of an image of class $c_\text{orig}$ such that the prediction logits of a target class $o_{c_{target}}(r)$ are maximized with minimal changes only. We therefore minimize: 
\begin{equation}
\label{eq:counterfactual}
L(r, \Delta r) = -o_{c_\text{target}}(r + \Delta r) + \lambda_1 o_{c_\text{orig}}(r + \Delta r) - \lambda_2 L_\text{ID}
\end{equation} 
where $\lambda_1$ and $\lambda_2$ are weighting coefficients, empirically set to $\lambda_1 = 0.6$, $\lambda_2 = 10$. 
$L_{\text{ID}}$ is an extra penalization term to preserve the identity of the object in the $W$-space:
\begin{equation}
L_{\text{ID}}(r, \Delta r) = \frac{f(r) f(r+\Delta r)}{\|f(r)\|_2 \|f({r+\Delta r})\|_2}
\end{equation}
The shift $\Delta r$ is optimized using gradient descent until the predicted class for image $G_s(f(r+\Delta r))$ is $c_\text{target}$ but at most for 2,000 steps.
\section{Experiments \& Results }
\label{sec:results}

In the following sections, we demonstrate the feasibility (Sec. \ref{sec:results_feasibility_LN}) and performance (Sec. \ref{sec:results_performance_LN}) of our proposed linking network. We then demonstrate how our method can aid our understanding of the learned representations in the hidden layers of a classifier (Sec. \ref{sec:single_units} - \ref{sec:decision_boundary}). In the main text, we report results with the ResNet-50 classifier (see Supplement for other classifier architectures).

\subsection{Similarities between $W$ and the representation space $R$}
\label{sec:results_feasibility_LN}

In this work, we introduce methods to interpret learned representations in CNNs trained on object classification by linking the representation space $R$ of the classifier to the $W$-space in StyleGAN-XL. Finding a simple (or even linear) mapping between $R$ to $W$ may be possible if the representations within the two spaces are sufficiently similar, which we study in the following. 

\begin{figure}[h]
  \centering
  \includegraphics[width=\textwidth]{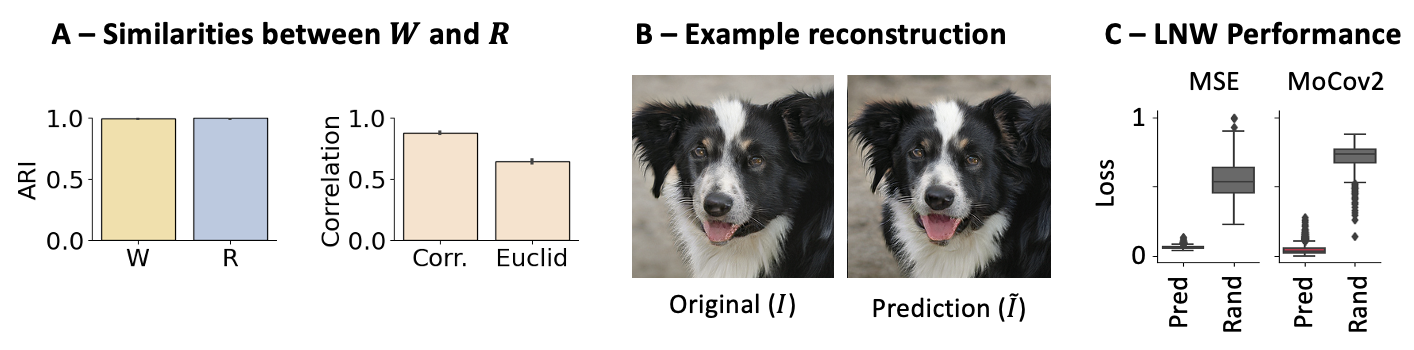}
  \caption{\textbf{Feasibility and performance of linking network.} \textbf{A)} High similarity between StyleGAN-XL's $W$-space and representation space $R$ in ResNet-50. Across 100 repetitions, 100 examples for five different ImageNet classes are sampled. \textbf{Left:} We fit a k-Means clustering ($k=5$, 20 initializations) on the selected examples and compute the Adjusted Rand Index (ARI) between the predicted clusters and the real class labels. \textbf{Right}: Learned representations in $W$ and $R$ are highly similar, shown here by high average correlations between flattened similarity (correlation) and dissimilarity (Euclidean distance) matrices of the selected examples computed for the two spaces. \textbf{B)} The trained linking network achieves high similarities between generated images ($I$) and images cycled through the linking network and the GAN ($\Tilde{I}$). \textbf{C)}  We quantify the performance of the linking network by the MSE between $w$ and $\Tilde{w}$ and by the perceptual image distance MoCov2 \cite{chen_improved_2020} between $I$ and $\Tilde{I}$.}
  \label{fig:fig3}
\end{figure}

First, since the penultimate layer of the classifier contains class-specific representations, we test if the $W$-space also separates classes. We test this using k-means clustering due to its simplicity, but other clustering or classification methods (supervised or unsupervised) could be used instead. Specifically, we repeatedly fit a k-Means clustering ($k=5$) to five randomly selected ImageNet classes and evaluate the performance using the Adjusted Rand Index (ARI) between the predicted clusters and the real class labels. We observe clustering by classes in both, the representation space and the $W$-space (Fig. \ref{fig:fig3}A left), indicating that $R$ and $W$ separate classes to a similar degree. 

Next, we use representation similarity analysis (RSA) to compare $R$ and $W$ on the representational level \cite{kriegeskorte_representational_2008}. We first compute similarity matrices (based on pairwise correlations) and dissimilarity matrices (based on the Euclidean distance) in $W$ and $R$ separately across sets of 5 randomly chosen classes (as above) and then compute the correlations between these (flattened) matrices in $W$ and $R$ revealing high similarities between the two spaces (Fig. \ref{fig:fig3}A right).

The robust clustering performance signifies a substantial degree of class demarcation within both spaces. The high representation similarity suggests a congruence in the representation of abstract concepts between $W$ and $R$. Together, these findings suggest that simple (linear) models may be adequate for establishing a functional linkage between these two spaces.

\subsection{Linking the representation space to $W$}
\label{sec:results_performance_LN}

The core of our method is a linear regression model that we train to link the representation space $R$ in the classifier to the $W$-space in the StyleGAN-XL (Fig. \ref{fig:fig2} red arrow, see Supplement for more complex (non-linear) linking networks). First, we generate 5,000 images per class for several classes, all of which are correctly classified. We then encode these images into the classifier and extract the activations in the representation layer $R$. We then train the linear regression on the pair of activations $r \in R$ and the corresponding $w \in W$, using 5,000 examples per class.

We assess the performance of this linking network in $W$ as well as in the image domain using a newly generated test set. We observe that after mapping an image for a full cycle, i.e., $w \to I \to r \to \Tilde{w} \to \Tilde{I}$, the mapped image is highly similar to the original image (Fig. \ref{fig:fig3}B). We quantify the loss in $W$ as the mean squared error (MSE) between the original $w$ and the cycled prediction $\Tilde{w}$ and observe a high performance (loss values close to 0) that is significantly better than that obtained for randomly selected images (Fig. \ref{fig:fig3}C left). Moreover, we obtain consistent results for a comparison within the image domain using the perceptual distance measure MoCov2 \cite{chen_improved_2020} (Fig. \ref{fig:fig3}C right, see Supplementary Fig. S1 for other image similarity metrics).

\subsection{Conceptualizing single-unit representations}
\label{sec:single_units}
\begin{figure}[t]
  \centering
  \includegraphics[width=\textwidth]{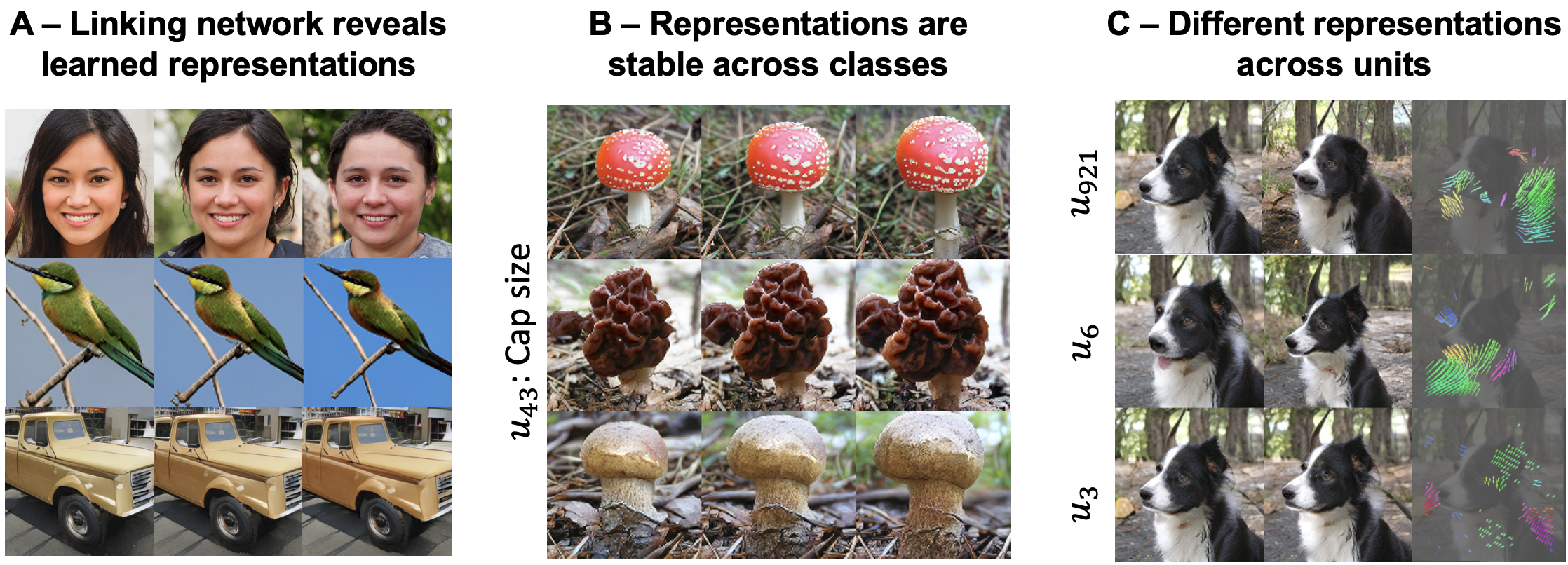}
  \caption{\textbf{Automatically revealed abstract concepts encoded in individual units.} We tune the activation of individual units in $R$ and visualize the results. We observe abstract concepts to be encoded in single units, such as gender or color (\textbf{A}), and to be stable across different classes, shown here for the cap sizes of different fungi (\textbf{B}). \textbf{C}) Different units encode different concepts that can be visualized with our unsupervised tracking method (vector fields).}
  \label{fig:fig4}
\end{figure}

In this section, we demonstrate how our method can be used to effectively visualize abstract concepts encoded in individual units in the classifier, and to quantify systematically such representations across large numbers of units. For a given input image and unit, we incrementally alter its activation, a process that yields a sequence of images, enabling a visual inspection of variations in salient features encoded in that unit. Our unsupervised method additionally visualizes these features in a vector field (Fig. \ref{fig:fig4}). As a classifier's hidden layers contain large numbers of units (here 2,048), such that visual inspection is impractical, we propose an automatic analysis pipeline allowing us to examine  \textit{all} units in $R$. Specifically, we use semantic image segmentation to evaluate changes in area, luminance, entropy, eccentricity, and rotation angle of each segmentation label. We find individual units that represent specific features such as gender or color (see Fig. \ref{fig:fig4}A). These single-unit representations can be robust across classes (Fig. \ref{fig:fig4}B) and differ between units (Fig. \ref{fig:fig4}C). Note that our method also reveals relevant features when the classifier and the GAN were trained on similar yet different datasets (see Supplementary Fig. S4).

\begin{figure*}
  \centering
  \includegraphics[width= \textwidth]{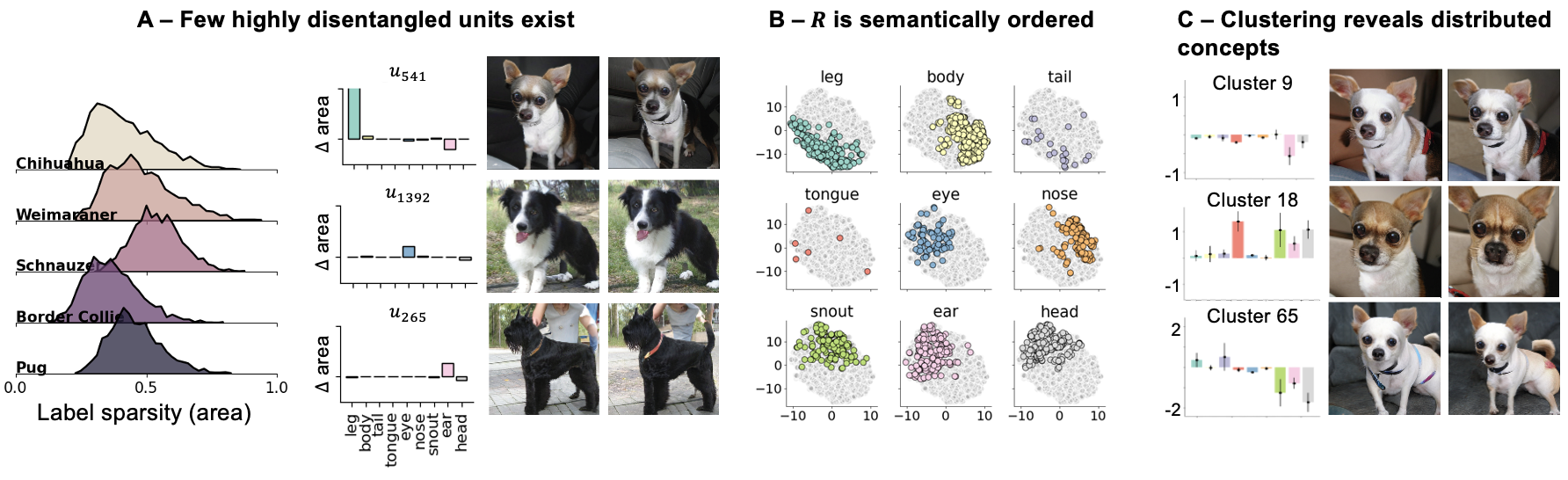}
  \caption{\textbf{Overview of features represented by individual units for the 2,048 units in the hidden layer of the ResNet-50 classifier.} \textbf{A}) Left: We compute the label sparsity of each quantification metric (here area, see Supplement for other metrics) across all units for 100 test seeds. Different classes exhibit different levels of sparsity. Right: Highly sparse units reveal disentangled representations of concepts such as long legs, larger eyes, or longer ears. \textbf{B}) $R$ is semantically ordered. We encode all changes in area induced by single-unit perturbations into a low-dimensional space using tSNE and color units by the label with the strongest change. Regional overlap between labels indicates interdependent representations of these concepts; as observed for snout, ear, and head, but not for legs and body. \textbf{C}) Hierarchical clustering of the label vectors reveals clusters representing disentangled concepts (cluster 9) as well as combinations of previously observed overlapping concepts (clusters 18 and 65).}
  \label{fig:fig5}
\end{figure*}

Next, to test whether $R$ contains disentangled feature representations, we quantify each unit's label sparsity (Eq. \ref{eq:sparsity}). We now focus on different dog classes, as dogs share many features but also show a high variability such as different fur colors, or ear shapes. We perturb each unit separately and compute the label sparsity from the median change in each label (e.g., legs, body, etc.; Eq. \ref{eq:sparsity}) across 100 example images. We find long tail distributions of label sparsity, with few, highly sparse units, that vary between classes (Fig. \ref{fig:fig5}A left), showing highly disentangled representation for features such as legs or ears (Fig. \ref{fig:fig5}A right). Next, asking whether $R$ exhibits a semantic order, we visualize single-unit representations using tSNE (Fig. \ref{fig:fig5}B). We find some labels to occupy specific regions in the representational space, with interdependence (overlap) between several labels (e.g. snout, ear, and head). Clustering the semantic concepts reveals disentangled (Fig. \ref{fig:fig5}C, cluster 9) as well as entangled representations (Fig. \ref{fig:fig5}C, clusters 18 and 65) suggesting superposition of several labels.

\begin{figure*}
  \centering
  \includegraphics[width=.9\textwidth]{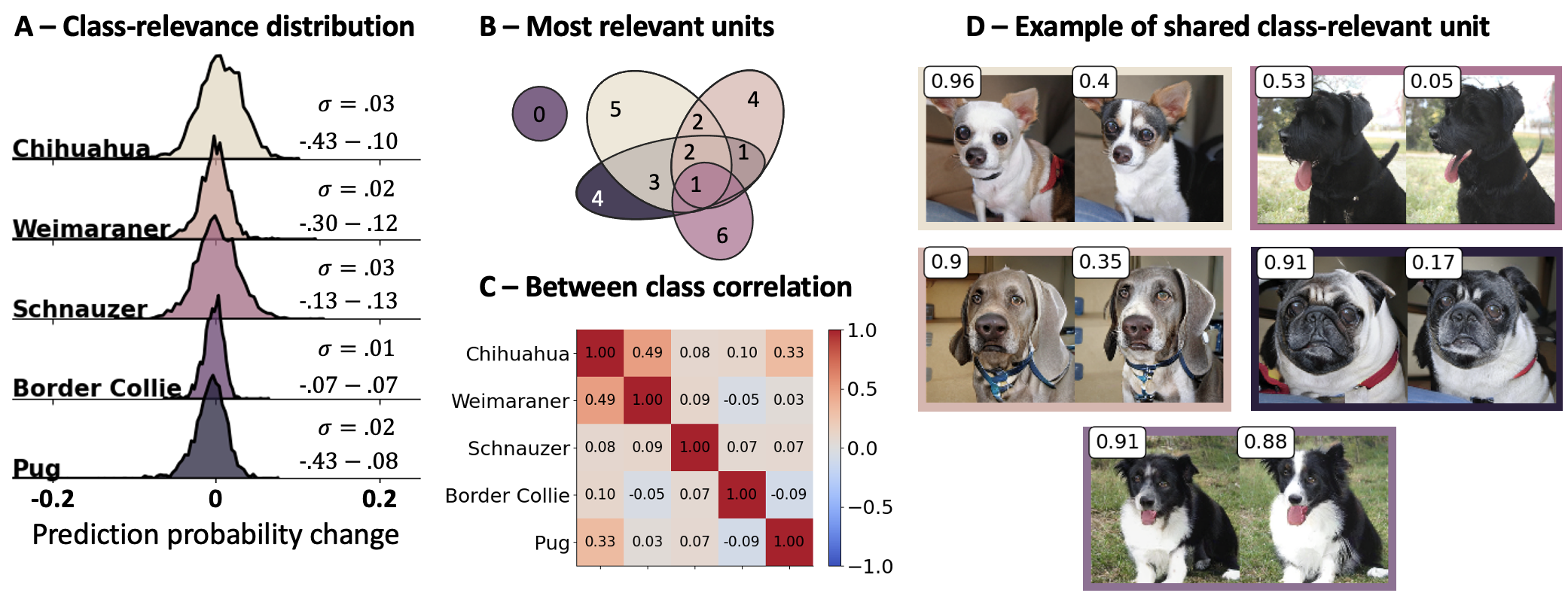}
  \caption{\textbf{Single units encode class-relevant representations.} We analyze the effect of single-unit perturbations on the classifier's prediction probabilities. \textbf{A}) We depict the distributions of the change in prediction probability (softmax) for five example classes averaged across 100 test seeds per class. \textbf{B}) Class-relevant units can be shared between classes. The Venn diagram shows the number of highly class-relevant units (average change in probability higher than 0.15) in each class. Further, our method can uncover robust classes that are not affected by single-unit perturbations (here e.g., Border Collie). \textbf{C}) Correlating the changes in the classifier's prediction over all units reveals similarities in class encodings. Whereas Chihuahua, Weimaraner, and Pug rely on similar representations (high correlations), Schnauzer and Border Collie appear to be encoded differently (low correlations). \textbf{D}) Varying the activation of the most relevant unit in B--which strongly affects prediction probabilities (upper left number) of all classes but the Border Collie (bottom row)--reveals human-interpretable features.}
  \label{fig:fig6}
\end{figure*}

\subsection{Class relevance of single units}
\label{sec:class_relevance}
Next, to systematically reveal the influence of single units on the classifier's prediction, we perturb each unit individually, visualize the new image, and extract the new prediction probability from the classifier given this image. We find that several units have a discernible effect on the prediction (Fig. \ref{fig:fig6}A). Our systematic quantification of the entire representation space allows us to identify class-relevant units (average change in probability greater than 0.15) that are unique to single classes, as well as a few units that are relevant to several classes (Fig. \ref{fig:fig6}B). Further, correlating the changes in prediction probability associated with each unit across classes reveals clusters of classes suggesting similar representational manifolds (e.g. between Chihuahua, Weimaraner, and Pug; Fig. \ref{fig:fig6}C). Additionally, visualizing the representations encoded in the class-relevant unit shared between most example classes (Fig. \ref{fig:fig6}B) reveals that such units can encode human-interpretable semantic concepts such as color or the length of the snout and can even cause a change in the predicted class (Fig. \ref{fig:fig6}D, see the change in the classifier's probability of predicting the original class indicated in the upper left corner).

\subsection{Discovering the classifier's decision boundaries}
\label{sec:decision_boundary} 
\begin{figure}
  \centering
  \includegraphics[width=\textwidth]{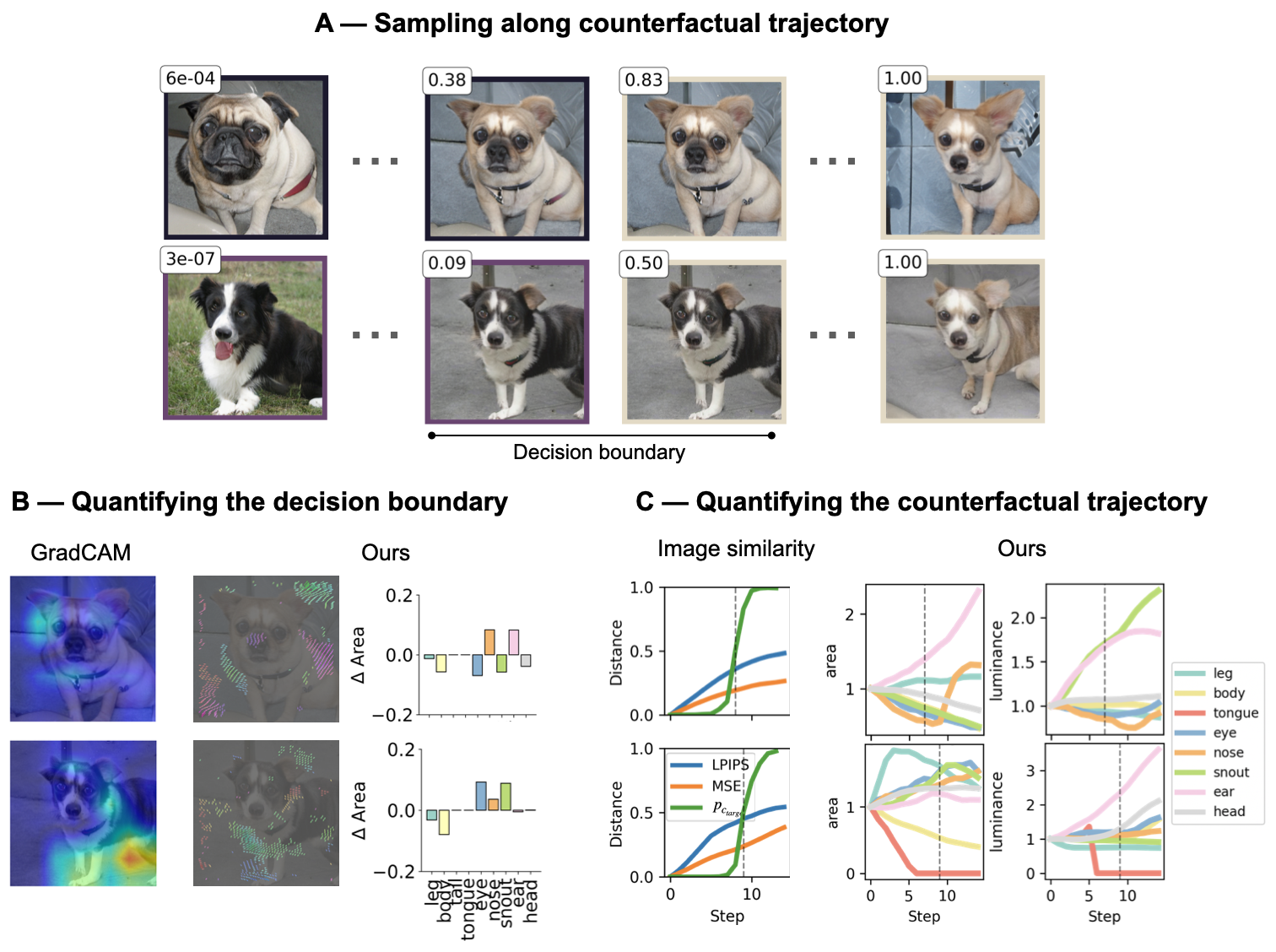}
  \caption{\textbf{Counterfactual trajectory between classes reveals the classifier's decision boundary.} Given an input image (upper row: Pug, bottom row: Border Collie, in A-C) we manipulate its activation $r \in R$ to produce a counterfactual example of a given target class (upper and bottom row: Chihuahua, in A-C). Based on our linking network, we visualize and quantify the representations along the counterfactual trajectory. \textbf{A)} Abrupt transitions in prediction probability at the decision boundary, with changes hardly visible to the human eye. For the two counterfactual examples, we show the original input image (left), the two images around the decision boundary (center), and the final image of the target class (right). The probability of the target class is indicated in the upper left corner. The predicted class for each image is coded by the frame-color. \textbf{B)} Quantifying the changes at the decision boundary reveals human-interpretable concept changes with our method (right), which previous methods, such as GradCAM \cite{selvaraju_grad-cam_2020} could resolve (left). \textbf{C)} Left: Common similarity metrics (MSE, LPIPS) fail to capture the abrupt transition in the classifier's prediction probability ($p_{c_\text{target}}$, green). Right: Our method reveals comprehensive trajectories to interpret a classifier's decision boundary across several labels and metrics.
  The decision boundary is indicated by the vertical dashed line. All metrics along the trajectory are computed using the original image as a reference and normalized.} 
  \label{fig:fig7}
\end{figure}

In Sec. \ref{sec:single_units} we show that our method can reveal learned representations in single units. However, our method can also be extended to visualize and quantify representations encoded in \textit{distributed} activations in $R$. Specifically, we analyze the representations that change across a classifier's decision boundary. For an image of a given class, we linearly manipulate the activation $r \in R$ to shift the prediction probability towards a target class (see Eq. \ref{eq:counterfactual}). We refer to the point in $R$ at which the predicted class changes as the decision boundary. Our proposed method allows to generate human interpretable visualizations of the representations along such counterfactual directions, showing that images at the decision boundary are visually indistinguishable, despite rapid changes in the prediction probability (Fig. \ref{fig:fig7}A, see probability insets). Using our quantification method, we can zoom in on specific concepts (ear, legs, etc.) and pinpoint how their representation changes across the decision boundary (Fig. \ref{fig:fig7}B right). Such level of specificity is not reached in previous models, including GradCAM \cite{selvaraju_grad-cam_2020}, which may emphasize less interpretable properties, such as features of the background (Fig. \ref{fig:fig7}B left). Along the counterfactual trajectory, common image similarity metrics (MSE, LPIPS) show smooth trajectories across the decision boundary (Fig. \ref{fig:fig7}C left). In contrast, our quantification pipeline draws a more comprehensive picture of the decision-relevant features learned by the classifier (Fig. \ref{fig:fig7}C right). For example, between Pug and Chihuahua, we observe a continuous increase in the area of the ears while the luminance saturates after the decision boundary (Fig. \ref{fig:fig7}C right, top row, pink lines). Between the Border Collie and Chihuahua, the luminance of the head, eyes, and body increases around the decision boundary (Fig. \ref{fig:fig7}C right, bottom row, gray and orange lines). Our method hence offers the opportunity to identify features (ir-)relevant to a classifier's decision.

\section{Discussion and Limitations}
\label{sec:discussion}
We introduce a method that leverages the semantic structure in a pre-trained GAN, and thus does not require intensive (re-)training as often suggested by previous work (e.g., \cite{lang_explaining_2021, casanova_instance-conditioned_2021, bordes_high_2022}). 
Our approach offers a computationally inexpensive method for analyzing several different classifiers (trained on similar datasets as the GAN), as training the linking network is fast. Our method even generalizes to models trained on similar but non-overlapping datasets (e.g., different face datasets). Currently, our method does not generalize to ViTs, where regional representations are encoded, as the $W$-space is not regionally disentangled. 

Our analysis pipeline uses a few-shot image segmentation model, allowing us to identify fine-grained features, while minimizing time-intensive image labeling. Note that, we use few-shot segmentation as currently, no large-scale segmentation dataset exists that contains as much feature detail. A network trained with more data could further improve the segmentation masks. However, averaging across many generated images, as done here, diminishes errors due to imprecise segmentation masks. Note that, despite being trained only on a limited set of classes (e.g, some dog breads), the image segmentation model generalizes to other, similar classes (Supplementary Fig. S6), which drastically reduces the amount of required labeled data. 

So far, our analyses use only generated images, as current generative models still cannot fully capture the diversity of real data. However, we assume that many of the features that a GAN can generate are also relevant features encoded in CNN classifiers. Note that despite this limitation, our approach can resolve fine details (e.g., eyes). We expect the development of more accurate generative models in the future will make applications of our approach to real images more robust. Despite not yet being readily applicable to real images, we believe our work introduces a novel type of approach to enable an unprecedented view of the learned features represented in hidden layers of the CNNs. 

Our analyses focus on highly similar classes that share general large-scale features (e.g., general body shape) but differ along fine-grained features. It is possible to train the linking network with more (diverse) classes. However, this would shift the emphasis of the revealed features to a more macro-scale level. 

In contrast to previous studies, we here analyze all individual dimensions of the representation space and compute comprehensive summary statistics. To the best of our knowledge, no benchmark has been proposed that can quantify abstract representations encoded in single units of CNNs. 
We believe that our method can open several avenues for studying how the representational geometry (e.g. feature (dis-)entanglement or sparsity) affects a model's performance and its robustness to adversarial attacks. In the future, insights about the complete representational space may inspire model architectures or training strategies that constrain the representational geometry. Additionally, extending our approach to other convolutional layers could reveal insights into features represented at different hierarchies in the classifier. Also, using our method to find difficult-to-classify images, as in \cite{prabhu_lance_2023}, appears to be an interesting future direction.





\section{Conclusion}
\label{sec:conclusion}

We introduce a simple, yet effective tool to visualize and systematically analyze the representations learned in a classifier. Our approach allows us to study abstract concepts encoded in individual units, as well as representations distributed over many units. The proposed automatic quantification pipeline provides a tool to interpret thousands of units simultaneously, demonstrating that even single units can encode meaningful, partially disentangled features carrying relevant class information, and revealing concepts that change across the decision boundary. We believe that our methods can provide novel insights into the representation of abstract concepts in the hidden layers of classifiers and thereby aid the introduction of such models in various real-world applications.

\section*{Acknowledgements}
The research leading to these results has received funding from the Deutsche Forschungsgemeinschaft (DFG, German Research Foundation) – 414985841 GRK 2566 "iMOL" (MW, POV, and MK), from the German Research Foundation - DFG Research Unit FOR 5368 ARENA (MK), and from SPP 2041 "Computational Connectomics" (POV and MK).


%
%

\clearpage
\bibliographystyle{splncs04}
\bibliography{references}

\end{document}